# 3M-TRANSFORMER: A MULTI-STAGE MULTI-STREAM MULTIMODAL TRANSFORMER FOR EMBODIED TURN-TAKING PREDICTION

*Mehdi Fatan, Emanuele Mincato, Dimitra Pintzou, Mariella Dimiccoli* *

Institut de Robòtica i Informàtica Industrial, CSIC-UPC, Barcelona, Spain

## ABSTRACT

Predicting turn-taking in multiparty conversations has many practical applications in human-computer/robot interaction. However, the complexity of human communication makes it a challenging task. Recent advances have shown that synchronous multi-perspective egocentric data can significantly improve turn-taking prediction compared to asynchronous, single-perspective transcriptions. Building on this research, we propose a new multimodal transformer-based architecture for predicting turn-taking in embodied, synchronized multi-perspective data. Our experimental results on the recently introduced EgoCom dataset show a substantial performance improvement of up to 14.01% on average compared to existing baselines and alternative transformer-based approaches. The source code, and the pre-trained models of our 3M-Transformer will be available upon acceptance [1].

## 1. INTRODUCTION

Over the last decade, there has been a growing interest in the automatic analysis of conversational data with the aim of improving our understanding of human-human communication and multimodal signaling of social interactions. Turn-taking prediction, understood as the task of predicting who is going to talk seconds ahead, is a fundamental task for conversational systems, with numerous human-centered applications, such as early diagnosis and intervention for communication disorders like autism [8], conversational systems [23], human-robot communications [21] to name but a few. Early approaches focused mainly on verbal communication, with audio cues being the primary factor for determining turn shifts. However, it is widely acknowledged that not only verbal indications but also vocal and visual information are essential for turn-taking prediction. In natural conversational settings, interlocutors use a variety of social signals, including mutual gaze, head position, facial expressions, breathing, and hand gesticulation, to communicate their speaking intentions to their partner [6]. Multimodal approaches for turn-taking prediction have traditionally used a third-person perspective for visual and audio data. However, human intelligence, including the ability to communicate effectively, evolved with sensory input from the egocentric, first-person perspective. Recently, the introduction of the Egocentric Communications dataset (EgoCom) [16], a multi-perspective and multimodal dataset, allowed to show the importance of synchronized multi-perspective embodied data to improve the accuracy and robustness of turn-taking prediction. In this paper, we propose a multimodal transformer-based architecture for turn-taking prediction that achieves impressive performance in synchronized multi-perspective conversational data beating state of the art multimodal transformer architectures by 14.01 % on average. These results make a huge step towards the dream of effective conversational systems in real world applications.

## 2. RELATED WORK

**Turn-taking and next speaker prediction.** The problem of turn-taking prediction has originally been addressed in the field of speech processing with the goal of building dialog systems able to avoid utterance collisions in multiparty meetings. Early systems could make a prediction only once the speaker was done, while models able to make future predictions continuously were introduced only later. To detect a potential turn shift after a segment of speech, typical cues considered were small gaps of silence in between speakers, semantic and lexicon-syntactic features and prosody [18, 23, 4]. With the widespread use of neural networks, Long Short-Term Memory (LSTM) models have assisted turn-taking tasks because of their abilities to capture temporal interactions [19].

As modeling turn-taking largely depend on nonverbal communication, gaze [5], head pose [1] and gesticulation were used not only to indicate a turn-taking point but also to clarify the addressee selection. Recent studies have raised the challenge of combining different modalities in turn-taking. Recent work proposed a real-time systems that predict turn-taking from acoustic features beside lexical information with a LSTM network [11], which includes several different types

*This work has been supported by project PID2019-110977GA-I00 funded by MCIN/ AEI /10.13039/501100011033 and by the "European Union NextGenerationEU/ PRTR", as well as by grant RYC-2017-22563 funded by MCIN/ AEI /10.13039/501100011033.

[1] `https://github.com/mehdifatan/Egocom-IRI-UPC`

of dialogue sessions. Later, a real-time generalized model was compared to a scenario-specific model, using both linguistic and acoustic features, and showed that generalized models works better in unstructured and informal conversations [7]. Also, lexical and prosody cues were associated in an RNN model to estimate the timing of a turn-taking and classify each utterance [10]. Parallel works focused on gaze transition patterns and the timing structure of eye contact between a speaker and a listener near the end of the speaker's utterance[5]. Recently, [13] predict the next speaker leveraging nonverbal features by using training data with rich combinations of participants. Most multimodal studies including vision have used a third-person visual perspective. However, this approach tends to lose part of the contribution of eyes' gaze, facial expressions, head pose and gestures of some of the participants, which are easily occluded in a third view-perspective. [9] showed the relevance of modeling eye activity from a wearable device. Very recently, the introduction of the EgoComm dataset [16] a multimodal, multi-view egocentric perspective datasets, demonstrates the benefit of such data capture setup for the task turn-taking prediction. We build on these results to validate our model.

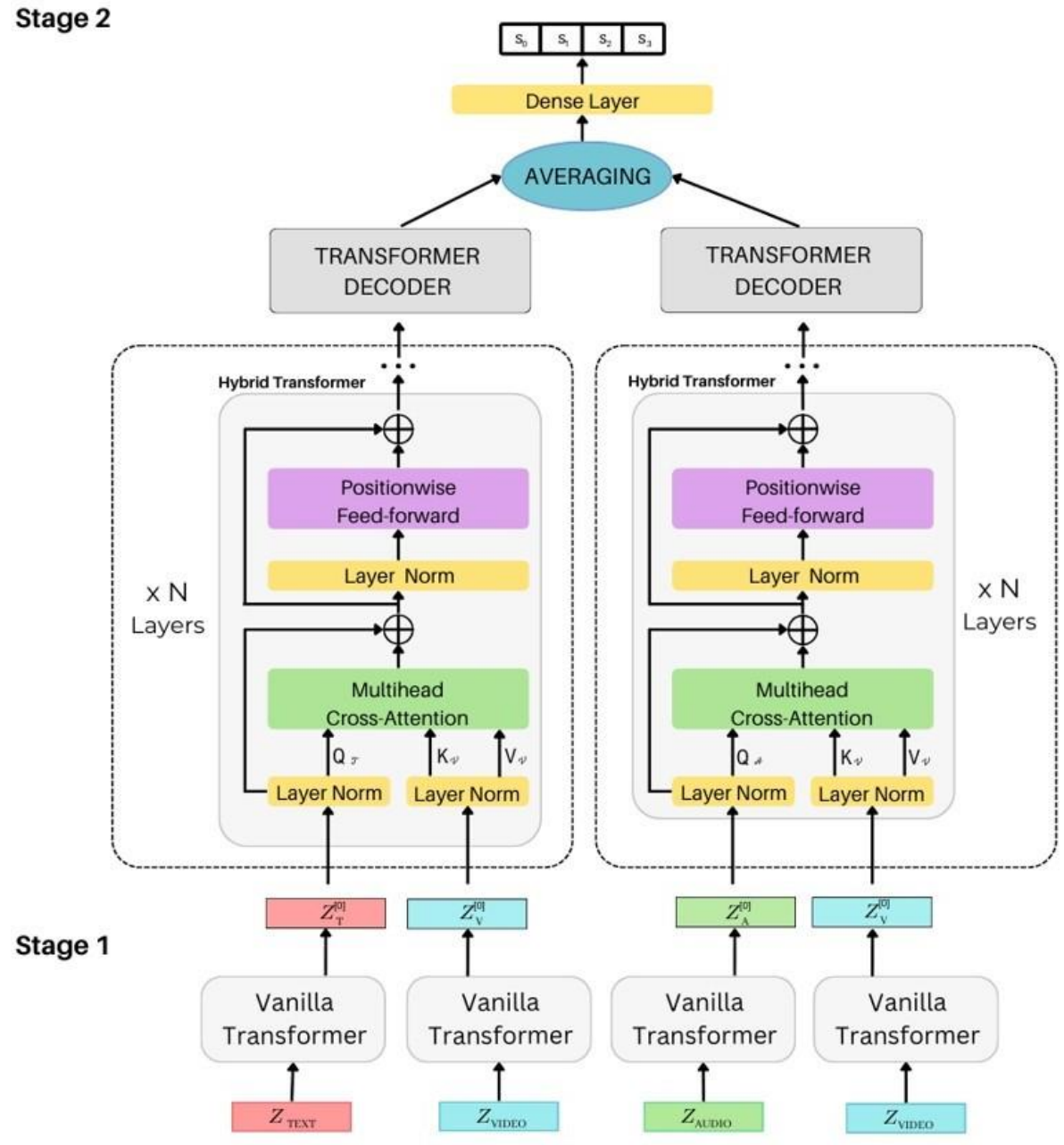

**Fig. 1**. Our 3M-transformer architecture: in the first stage the initial features are given as input to three different vanilla transformers. Their decoder's outputs are combined and used as input to a two-stream Hybrid Transformer. The first and the second stream use as Query the text features ($Z_T$ ) and the audio features ($Z_A$), respectively, and the video features ($V_S$) as Key and Value. The outputs of each stream are soft-averaged and feed to a projection head for speaker classification.

**Multimodal transformer-based architectures.** Transformer-based models, originally developed in the field of NLP [25] have proven to be effective for a variety of tasks where data can be approximated by a sequence of discrete units, such as video classification [2] and audio classification [22]. More recently, multimodal transformer-based architectures have shown great promise in handling multimodal data for a variety of tasks. From generic pre-trained models [3], like leveraging video and text, to task-specific architectures such as emotion recognition [17], audio-visual voice separation [14] and crowd counting [20] among others.

## 3. PROPOSED METHODOLOGY

We propose a cross-modal transformer-based architecture that casts turn-taking prediction as speaker classification task.
**Background.** The transformer model, as first proposed in [25], consists of a stack of encoder and decoder, with six identical layers.The capability of the network to capture sequence non-linearities lies mainly in the attention modules, that is built through a scaled dot-product on which it is based the Multi-Head Attention. One input sequence, called "query" (Q), is compared to the other input sequences, which are called "keys" (K) by a scaled dot product, scaled by the equal query and key $d_k$. Then the output is weighted with the values" (V) sequence input. This is described by the function:

$$Attention(Q, K, V) = Softmax\left(\frac{Q \cdot K^T}{\sqrt{d_k}}\right)V$$

Thanks to self-attention and encoder-decoder attention modules, the transformer network learns on which past position it needs to focus to predict a correct speaker.
**Network architecture.** We propose a multi-stage, multi-stream, multi-modal (3M)-transformer-based architecture. In the first stage, each modality (V,T,A) is processed independently by a transformer that takes as input a sequence of feature observations (past), and output a sequence of feature predictions . These unimodal predictions are further combined and refined in a second stage consisting of multi-stream hybrid transformer (HT) architecture where in each stream a different modality is used as query, while the key and value remain constant. The outputs of the two streams are finally concatenated through the average and a fully connected layer followed by softmax is used to produce the final prediction of the next speaker. In particular, the A→V and the T→V units take in audio and textual features respectively as the query and the visual features as the key and values.

## 4. EXPERIMENTS

### 4.1. Experimental setup

**Dataset.** To train, evaluate and test our model we used the EgoCom dataset [16], consisting of 28 different 20-30

minutes long natural conversations between three speakers. Synchronized videos are captured by the point of view of at least two speakers wearing video and audio recording glasses. All conversations are directed by a leading person who has the role of the host and cover different topics. The dataset is already split into train, validation and test sets with the percentages of 78%, 6% and 16%, respectively. It provides video and audio records as well as the text content of the conversation, ground truth transcriptions and the speaker labels.

**Features.** We used the features provided in [16]. The video embeddings are 2048-dimensional visual features corresponding to the last average pooling layer of R(2+1)D-101 model [24], pre-trained on Kinetics-400. The audio embeddings are 64-dimensional vectors extracted by using a speaker identification model trained on the Voxceleb [15]. The text embeddings are 300-dimensional features obtained by using the transcripts by FastText's Crawl[12]. For each second, 12 feature embeddings were created by an overlapping sliding window for each modality (text, video, audio). Thereafter, the features were collected to represent the past 4, 5, 10, 30 seconds for every modality and their combination.

**Evaluation protocol.** As we formulated the problem of turn-taking prediction as a classification task aimed at identifying the next speaker (no one, host or one of the two participants), we used the classification accuracy as evaluation metric.

**Baseline models.** In addition to the MLP model introduced in [16], we introduced two transformer-based baselines we named Early Fusion Transformer (EFT), and Late Fusion Transformer (LFT). EFT is a vanilla transformer that takes as input a sequence of multimodal features given by the concatenation of features corresponding to different modalities (text, audio, vision), covering a time interval in the past. LFT processes each modality separately by a vanilla transformer. Afterward, a soft ranking layer combines the output of transformers to produce the final probabilities of next speaker.

**Implementation details.** We predict the speaking label $t$ seconds in the future both without (i.e. likelihood) and with (i.e. posterior) inclusion of the current speaking label (i.e. prior) as input during training. In all experiments, the input are the pre-computed features corresponding to an interval of time in the past 4, 5, 10, 30 seconds, and the output is the prediction of the next speaker after 1, 3, 5, 10 seconds. There are 3 modalities, text (T), audio (A), video (V).The embedding size for all proposed transformer based models and the attention layer have been set to 512 which was enough for coding input feature vectors. The embedding method is a linear transform with trainable weights and bias and after which the positional encoding is applied. All the transformer models are in the standard form of encoder-decoder model with feed-forward layer size of 2048 neurons and 8 as number of heads for multi-head attention layers. The 6 sub layers has been used for encoder and decoder parts of all transformer models. Finally, we used a dropout value of 0.1 while, to train our architecture, we chose the Adam method as optimizer, with a learning rate of 0.01 and a weight decay of 1e-7.

### 4.2. Ablation study

In Table 1, we report the results of the ablation study we performed to validate the proposed 3M-Transformer architecture as well as the combination of modalities. In particular, we validated the multi-stage, the multi-stream, and the multi-modal design. $X \rightarrow Y$ as subindex means that features $X$ are given as Q, and features $Y$ as K and V. In table 1 results are reported in blocks, where each blocks (rows divided by horizontal lines), refers to a different structure of the architecture and each row within the block represents a different combination of Key, Value, Query of the input features. The first block is obtained considering the full architecture described in section 3. The second block (fourth row) is the case where the output features of the second stage are concatenated, instead of taking the average. This validates the choice of doing soft-averaging instead of the concatenation of output features before the classifier. The third block is the case where only one stream is considered in the second stage of the architecture, so here no type of merging of the output feature is needed. The results obtained in this way by using only two modalities are very similar to those obtained by using three modalities. We observed a similar behaviour with EFT, LFT and MLP. Combining $A$ and $T$ results to be slightly the best choice. The fourth block instead is like one but here the difference is that the decoder is not implemented in the second stage of the architecture. Finally in the last block are reported the cases where one stage is omitted: in the penultimate row there are the results of the architecture without the second stage while in the last row the results obtained directly with the second stage. These results clearly validate the multi-stage approach.

### 4.3. Comparative results

Table 2 provides comparative results in terms of test accuracy of our proposed approach in comparison with the MLP method introduced in [16], the transformer-based baselines we introduced above (EFT and LFT), and two state of the art transformer-based cross-modal architectures [17], denoted as ATT and CROSS. These results show that all transformer-based architectures perform much better than the MLP method even in the single modality setting, and that the combination of two modalities already achieves performance similar to those obtained by using three modalities. Compared to methods in [17], which make accurate predictions only in the short-term, our 3M-Transformer is very accurate in both short time and long time, even using a few seconds of past. We observe also that all transformer-based models demonstrate superior performance in capturing and leveraging the past for future predictions, as they exhibit only a minor decrease in accuracy when trained with features averaged over a larger past/history with respect to the MLP model.

**Table 1**. Ablation study on the EgoComm dataset. All the results (Top-1 accuracy) are obtained considering the *Prior* = True

| Past(s) | 4 | | | | 5 | | | | 10 | | | | 30 | | | | Average |
|---|---|---|---|---|---|---|---|---|---|---|---|---|---|---|---|---|---|
| Future(s) | 1 | 3 | 5 | 10 | 1 | 3 | 5 | 10 | 1 | 3 | 5 | 10 | 1 | 3 | 5 | 10 | |
| 3M-Transformer$_{T\to V\,\|A\to V}$ | 93.62 | **95.73** | **95.63** | 94.72 | 95.13 | **95.07** | 94.66 | 94.80 | 94.98 | **94.59** | 93.76 | 93.42 | **95.20** | 94.43 | 94.35 | 92.14 | **94,51** |
| 3M-Transformer $_{V\to T\|A\to T}$ | **95.84** | 93.41 | 95.31 | 95.00 | **95.34** | 93.79 | 93.48 | 94.32 | **95.45** | 94.06 | 92.56 | 94.34 | 93.33 | 95.20 | 95.2 | 93.3 | 94,37 |
| 3M-Transformer $_{V\to A\|T\to A}$ | 92.99 | 90.20 | 95.35 | 94.57 | 92.65 | 94.89 | 94.15 | 95.12 | 94.77 | 93.16 | 92.98 | 94.42 | 94.81 | 91.67 | 94.60 | 93.29 | 93,72 |
| 3M-Transformer $_{T\to V\,\|A\to V}$ | 93.85 | 95.33 | 93.37 | 93.92 | 93.28 | 93.93 | 92.67 | 94.18 | 92.28 | 93.30 | 93.90 | 94.99 | 92.47 | 93.55 | 92.94 | 93.22 | 93,66 |
| 3M-Transformer $_{V\to T}$ | 93.97 | 95.09 | 91.13 | 91.98 | 94.33 | 93.92 | 93.05 | 94.36 | 94.84 | 93.00 | 91.62 | 93.96 | 91.91 | 93.41 | 94.78 | 88.10 | 93,09 |
| 3M-Transformer $_{V\to A}$ | 94.32 | 92.30 | 95.06 | 94.56 | 94.98 | 94.57 | **95.32** | **95.16** | 94.68 | 91.65 | **95.57** | 94.88 | 91.91 | 93.67 | 93.19 | 94.20 | 94.12 |
| 3M-Transformer $_{T\to V}$ | 94.03 | 95.67 | 93.69 | 95.18 | 94.29 | 93.39 | 90.93 | 93.85 | 94.48 | 91.56 | 94.83 | **95.05** | 94.63 | 94.14 | 93.92 | **94.63** | 94.12 |
| 3M-Transformer $_{T\to A}$ | 94.52 | 95.59 | 95.94 | 89.86 | 93.85 | 93.04 | 92.65 | 93.69 | 95.01 | 90.52 | 92.33 | 91.79 | 92.79 | 94.01 | 93.43 | 90.61 | 93.10 |
| 3M-Transformer $_{A\to V}$ | 93.97 | 94.79 | 92.79 | 94.05 | 94.16 | 93.14 | 92.90 | 94.25 | 95.03 | 90.83 | 94.98 | 92.48 | 92.41 | 94.72 | **95.22** | 92.88 | 93.66 |
| 3M-Transformer $_{A\to T}$ | 94.79 | 93.97 | 95.59 | **95.37** | 94.95 | 94.54 | 93.34 | 93.27 | 94.85 | 91.21 | 94.63 | 94.46 | 92.81 | **95.50** | 94.32 | 93.63 | 94.20 |
| 3M-Transformer$_{T\to V\,\|A\to V}$ | 79.08 | 79.42 | 79.53 | 76.53 | 77.94 | 74.96 | 79.42 | 78.88 | 78.67 | 77.75 | 76.19 | 76.83 | 75.80 | 75.91 | 77.77 | 77.55 | 77.63 |
| 3M-Transformer$_{V\to T\|A\to T}$ | 79.28 | 78.13 | 79.82 | 73.08 | 77.96 | 76.28 | 77.35 | 75.07 | 80.11 | 78.37 | 79.67 | 78.04 | 78.38 | 79.92 | 79.30 | 76.17 | 77.93 |
| 3M-Transformer$_{V\to A\|T\to A}$ | 78.97 | 79.26 | 78.67 | 76.19 | 77.97 | 74.78 | 76.69 | 76.00 | 79.75 | 78.87 | 78.71 | 75.19 | 78.31 | 79.62 | 77.17 | 75.87 | 77.62 |
| 3M-Transformer $_{W/o\ 1^{st}part}$ | 60.79 | 61.30 | 57.81 | 55.57 | 64.16 | 59.02 | 58.31 | 57.12 | 63.29 | 59.26 | 56.98 | 55.99 | 65.36 | 61.37 | 55.98 | 55.65 | 55.99 |
| 3M-Transformer $_{W/o\ 2^{st}part}$ | 59.43 | 57.97 | 55.15 | 50.57 | 62.85 | 58.80 | 57.72 | 57.11 | 66.48 | 60.40 | 57.85 | 55.83 | 64.76 | 59.71 | 58.05 | 56.54 | 58.70 |

**Table 2**. Comparative results in terms of Top-1 accuracy on the EgoCom dataset.

| (data used for training) | | Past(s) | 4 | | | | 5 | | | | 10 | | | | 30 | | | |
|---|---|---|---|---|---|---|---|---|---|---|---|---|---|---|---|---|---|---|
| Use Prior | Modalities | Future(s) | 1 | 3 | 5 | 10 | 1 | 3 | 5 | 10 | 1 | 3 | 5 | 10 | 1 | 3 | 5 | 10 |
| False | T | EFT/LFT | **63.71** | **57.28** | **55.07** | **49.29** | **61.11** | **57.95** | **56.22** | **55.52** | **61.67** | **56.25** | **55.59** | **56.10** | **63.21** | **56.74** | **53.79** | **55.70** |
| *(likelihood)* | | MLP | 53.5 | 47.8 | 47.3 | 45.9 | 51.6 | 47.5 | 45.6 | 44.9 | 48.6 | 45.1 | 44.8 | 45.1 | 44.8 | 44.0 | 44.8 | 44.7 |
| | V | EFT/LFT | **60.55** | **55.59** | **53.44** | **54.92** | **55.54** | **56.20** | **56.64** | **57.03** | **55.56** | **55.77** | **56.79** | **56.86** | **56.15** | **56.38** | **56.33** | **56.17** |
| | | MLP | 44.6 | 45.3 | 43.9 | 42.2 | 45.4 | 45.0 | 44.9 | 44.9 | 44.4 | 45.8 | 45.1 | 44.4 | 43.2 | 44.8 | 44.8 | 44.6 |
| | A | EFT/LFT | **65.16** | **57.54** | **55.43** | **43.71** | **59.29** | **56.09** | **55.50** | **57.19** | **64.09** | **53.48** | **56.93** | **57.32** | **62.57** | **56.82** | **56.21** | **56.63** |
| | | MLP | 53.7 | 48.6 | 47.8 | 46.3 | 52.6 | 48.0 | 46.3 | 45.5 | 49.9 | 46.4 | 46.3 | 45.2 | 45.6 | 43.4 | 44.8 | 43.6 |
| | T+V | LFT | 56.69 | 57.41 | **56.69** | **55.11** | **58.48** | **57.80** | **60.30** | 57.12 | **57.18** | 56.67 | 56.51 | **56.78** | **56.13** | 55.64 | 55.86 | **56.81** |
| | | EFT | **60.31** | **57.43** | 56.11 | 47.32 | 57.69 | 56.90 | 56.86 | **57.30** | 55.93 | **56.75** | **56.80** | 56.62 | 54.81 | **55.66** | **55.95** | 56.66 |
| | | MLP | 48.3 | 44.9 | 43.7 | 44.5 | 48.5 | 45.3 | 45.4 | 44.5 | 45.1 | 44.2 | 45.1 | 44.5 | 40.1 | 44.5 | 44.7 | 44.5 |
| | T+A | LFT | 61.10 | 56.69 | **55.43** | **51.38** | 59.15 | **57.58** | **55.62** | **57.35** | **57.75** | **57.41** | **55.99** | **57.34** | **56.71** | **56.54** | **56.86** | **57.28** |
| | | EFT | **61.35** | **59.73** | 54.93 | 47.46 | **59.39** | 55.52 | 55.43 | 55.81 | 55.54 | 55.37 | 55.67 | 56.21 | 55.37 | 55.31 | 54.82 | 56.23 |
| | | MLP | 54.5 | 48.3 | 47.7 | 46.0 | 53.5 | 48.2 | 46.2 | 45.3 | 49.9 | 47.1 | 45.7 | 44.7 | 44.0 | 44.4 | 44.9 | 44.7 |
| | V+A | LFT | 60.16 | **59.15** | 57.41 | 55.34 | **58.37** | **58.18** | **57.60** | 56.35 | **57.22** | **56.59** | 56.44 | 55.84 | **56.82** | **56.98** | **55.92** | 56.42 |
| | | EFT | **64.15** | 58.69 | **57.86** | **49.55** | 58.28 | 57.94 | 56.43 | **56.57** | 56.54 | 55.51 | **56.45** | **56.53** | 56.01 | 56.00 | 55.25 | **56.70** |
| | | MLP | 51.6 | 47.2 | 46.9 | 42.1 | 52.2 | 46.9 | 45.5 | 44.9 | 46.9 | 45.0 | 45.2 | 44.9 | 44.0 | 44.3 | 44.9 | 44.8 |
| | T+V+A | LFT | 60.93 | 59.89 | 58.54 | 55.89 | 59.66 | 58.93 | 58.09 | 56.73 | 56.18 | 57.28 | 57.16 | 56.88 | 56.80 | 56.91 | 55.37 | 57.23 |
| | | EFT | 64.75 | 58.58 | 56.15 | 50.12 | 58.04 | 58.01 | 56.18 | 56.25 | 54.22 | 54.77 | 55.92 | 56.81 | 56.48 | 56.41 | 56.37 | 56.51 |
| | | MLP | 53.0 | 47.0 | 46.8 | 44.0 | 53.1 | 46.6 | 45.6 | 44.9 | 47.4 | 45.7 | 45.7 | 44.9 | 42.7 | 43.3 | 43.6 | 44.4 |
| | | ATT[17] | 66.91 | 63.52 | 58.39 | 54.36 | 60.69 | 58.49 | 54.43 | 50.93 | 53.60 | 50.84 | 48.41 | 47.06 | 36.80 | 40.46 | 36.53 | 39.31 |
| | | CROSS[17] | 63.52 | 61.03 | 53.88 | 45.33 | 64.84 | 55.50 | 53.42 | 45.07 | 58.86 | 53.28 | 50.92 | 43.42 | 48.35 | 44.25 | 45.44 | 44.28 |
| | | **Our** | **95.**53 | **95.**85 | **94.**64 | **95.**17 | **94.**25 | **94.**51 | **94.**60 | **93.**85 | **94.**13 | **94.**19 | **95.**25 | **92.**65 | **93.**56 | **92.**43 | **93.**91 | **93.**29 |
| True | T | EFT/LFT | **65.77** | **60.73** | **55.73** | **50.72** | **63.98** | **59.20** | **56.75** | **55.91** | **65.84** | **58.61** | **55.46** | **56.20** | **66.22** | **60.56** | **55.61** | **54.73** |
| *(posterior)* | | MLP | 56.8 | 48.8 | 47.5 | 45.8 | 56.4 | 47.9 | 45.3 | 44.6 | 53.5 | 46.8 | 45.2 | 45.0 | 55.2 | 46.6 | 44.8 | 44.7 |
| | V | EFT/LFT | **59.69** | **56.44** | **55.64** | **55.92** | **59.54** | **54.34** | **56.90** | **56.86** | **57.42** | **56.55** | **56.40** | **57.07** | **61.11** | **54.88** | **56.13** | **56.61** |
| | | MLP | 48.2 | 45.3 | 42.7 | 44.3 | 51.8 | 45.6 | 45.8 | 45.8 | 45.8 | 45.7 | 44.7 | 44.8 | 50.7 | 45.7 | 44.7 | 44.7 |
| | A | EFT/LFT | **65.69** | **61.34** | **54.36** | **47.59** | **65.91** | **58.60** | **56.55** | **57.12** | **67.17** | **59.60** | **54.10** | **57.15** | **64.63** | **61.24** | **56.70** | **56.98** |
| | | MLP | 56.5 | 49.0 | 47.7 | 46.2 | 57.2 | 48.4 | 46.4 | 44.9 | 54.2 | 47.0 | 46.3 | 45.1 | 55.1 | 45.7 | 45.0 | 44.5 |
| | T+V | LFT | **65.12** | **60.61** | **58.01** | **55.48** | **62.30** | **57.50** | **56.64** | 56.78 | **63.78** | **58.32** | **56.68** | **57.00** | **63.18** | **57.93** | **56.61** | **56.97** |
| | | EFT | 59.50 | 57.94 | 56.57 | 54.66 | 58.32 | 57.03 | 56.12 | **56.82** | 56.82 | 55.88 | 55.52 | 56.82 | 57.56 | 55.32 | 55.28 | 56.42 |
| | | MLP | 52.1 | 46.8 | 44.4 | 44.4 | 53.2 | 46.1 | 45.8 | 44.8 | 47.9 | 45.6 | 44.6 | 44.9 | 49.8 | 46.2 | 43.1 | 44.7 |
| | T+A | LFT | **67.46** | **61.27** | **56.16** | 47.76 | **64.57** | **58.68** | **57.19** | **57.29** | **67.31** | **59.69** | **57.18** | **57.43** | **65.15** | **59.58** | **57.12** | **57.29** |
| | | EFT | 65.68 | 60.08 | 55.06 | **48.78** | 64.11 | 57.25 | 55.43 | 56.16 | 65.67 | 56.73 | 56.38 | 55.87 | 60.40 | 59.15 | 56.56 | 56.51 |
| | | MLP | 56.9 | 49.0 | 47.8 | 46.1 | 56.9 | 48.5 | 46.3 | 44.8 | 54.3 | 47.0 | 46.0 | 44.7 | 54.4 | 45.9 | 45.0 | 44.9 |
| | V+A | LFT | 63.55 | **60.57** | **58.56** | **55.08** | **60.50** | **57.79** | **57.48** | 56.66 | **61.61** | 56.94 | 56.86 | 56.01 | **62.22** | **56.93** | 55.90 | 56.33 |
| | | EFT | **66.21** | 59.32 | 57.59 | 47.39 | 59.84 | 57.52 | 57.30 | **57.11** | 56.65 | **57.03** | **57.29** | **57.04** | 58.29 | 56.30 | **56.59** | **56.37** |
| | | MLP | 53.5 | 47.4 | 46.8 | 45.8 | 54.3 | 47.7 | 46.8 | 45.1 | 49.4 | 46.7 | 44.6 | 44.6 | 50.7 | 44.6 | 44.1 | 44.7 |
| | T+V+A | LFT | 64.33 | 61.49 | 59.18 | 56.16 | 61.66 | 58.83 | 57.71 | 57.29 | 63.22 | 58.38 | 57.30 | 56.68 | 63.24 | 56.02 | 56.86 | 57.21 |
| | | EFT | 62.15 | 59.28 | 55.23 | 47.92 | 58.16 | 56.90 | 56.48 | 56.40 | 56.60 | 56.70 | 55.62 | 55.94 | 58.14 | 55.63 | 56.55 | 56.47 |
| | | MLP | 55.4 | 47.0 | 46.7 | 43.9 | 55.1 | 47.3 | 46.3 | 44.8 | 48.7 | 45.9 | 45.3 | 44.2 | 50.7 | 44.8 | 43.2 | 42.6 |
| | | ATT[17] | 90.98 | 91.68 | 92.40 | 89.06 | 79.86 | 80.83 | 81.46 | 78.37 | 68.07 | 65.34 | 63.33 | 62.04 | 42.10 | 42.05 | 41.38 | 40.88 |
| | | CROSS[17] | 86.54 | 73.18 | 68.57 | 44.68 | 86.21 | 82.85 | 54.96 | 44.91 | 88.61 | 81.03 | 70.69 | 44.95 | 87.20 | 80.28 | 68.00 | 44.63 |
| | | **Our** | **93.62** | **95.73** | **95.63** | **94.72** | **95.13** | **95.07** | **94.66** | **94.80** | **94.98** | **94.59** | **93.76** | **93.42** | **95.20** | **94.43** | **94.35** | **92.14** |

## 5. CONCLUSION

This paper introduces a transformer-based model for multimodal turn-taking prediction in embodied multi-perspective, multimodal data. We have shown that it achieves impressive results outperforming cross-modal transformer-based baselines and architectures. Our work is significant as it ensure very reliable predictions suited for real world applications on embodied multi-perspective data. Future work will focus on reducing the amount of annotated data required.